# Scalable High-Speed Lateral Control for Single-Body and Articulated Autonomous Vehicles

Aashish Shaju, Steve Southward, and Mehdi Ahmadian

Center for Vehicle Systems and Safety (CVeSS), Virginia Tech, Blacksburg, VA 24060, USA

**Keywords:** High-speed lateral control, Articulated vehicles, Clothoid-based controller, Adaptive lookahead, Path tracking, Scalable control framework.

## Abstract

This paper presents a scalable lateral control framework designed to enable robust path tracking for both single-body and articulated autonomous vehicles under high-speed conditions. Traditional lateral controllers often struggle with oscillations, instability, and reduced accuracy when applied to high-speed maneuvers or complex vehicle configurations such as tractor-trailers. To address these challenges, a clothoid-based lateral controller is extended with several high-speed adaptations: 1) an integrated tangential check and Fréchet distance method for optimized lookahead and oscillation mitigation; 2) real-time trajectory segment classification for dynamic lookahead search space adjustment; and 3) a dual-adaptive, rate-controlled lookahead mechanism responsive to both cross-track error and proactive segment analysis. These enhancements enable dynamic tuning of the controller's lookahead distance to improve stability and accuracy during transitions between straight and curved paths. To ensure applicability across different vehicle types, the framework incorporates a flexible tracking point selection scheme and a curvature-to-steering lookup table (LUT). This allows accurate tracking of key vehicle points, such as the tractor rear axle or the hitch point, without structural changes to the control logic. The proposed controller is validated in simulation using high-fidelity TruckSim® models across varied driving scenarios, including dual lane changes and winding road sections. Results demonstrate improved stability, reduced oscillations, enhanced tracking accuracy, effective lookahead adaptation, and maintained lateral acceleration within safe limits across a wide speed spectrum and vehicle configurations. This study highlights the potential of a unified lateral control architecture for diverse autonomous vehicle platforms and lays the groundwork for further development toward multi-trailer systems.

## 1. Introduction

The reliable operation of autonomous vehicles (AVs) in high-speed and complex traffic environments demands robust lateral control systems. Effective lateral control is crucial for stability and efficient navigation at highway speeds, where nonlinear vehicle dynamics and susceptibility to oscillations increase. These challenges are substantially amplified in articulated vehicles, such as commercial semi-tractor trailers, due to complex dynamic coupling, distinct component curvatures, and phenomena like rearward amplification, necessitating specialized control strategies. Thus, there is a clear need for lateral control frameworks that can adapt to diverse vehicle configurations while maintaining stability at high speeds. This paper addresses this need by presenting a scalable control framework, extending a clothoid-based path-tracking architecture[1] with enhancements for articulated vehicle applications.

Various lateral control strategies have been investigated. Geometric controllers like Pure Pursuit and Stanley are common but can suffer with fixed lookahead distances [2]. Model Predictive Control (MPC) offers robust performance [3] but can be computationally intensive, especially for articulated vehicles [4]. Sliding Mode Control (SMC) provides robustness to uncertainties [5]. A critical parameter in many controllers is the lookahead distance, where fixed values are often suboptimal; thus, adaptive schemes adjusting lookahead based on speed or curvature have been explored [6,7]. The control of articulated vehicles presents unique complexities, with approaches ranging from LQR [8] to blended kinematic/dynamic strategies [9] and advanced MPC [10]. While existing methods show progress, many are tailored to specific vehicle types or computationally demanding, highlighting a need for broadly scalable lateral control frameworks for diverse configurations at high speeds. This paper presents such a framework and validates it through extensive simulation. Key contributions include a dynamic lookahead adaptation scheme, flexible tracking point selection, and integration with lookup table-based curvature control to enable seamless scalability.

This paper is structured as follows: Section 2 details the control methodology. Section 3 presents simulation results and discussions. Section 4 concludes the paper and discusses future work.

# 2. Scalable High-Speed Lateral Control Framework

This section details the development of a lateral control framework designed for effective high-speed operation across both single-body and articulated autonomous vehicles. The main approach involves improving the selection of the lookahead distance and introducing specific modifications to handle the effects of multi-body dynamics, with the goal of maintaining stable and accurate path tracking.

## 2.1 Overview of the Clothoid Controller and the Pivotal Role of Lookahead Distance

The foundation of the proposed controller is a clothoid-based path tracking algorithm, previously detailed in [1], which iteratively computes an optimal clothoid segment connecting the current vehicle pose to a forward point on the reference path. The clothoid, defined by a curvature profile κ(s), is selected to minimize deviation over a specified lookahead distance R, and converted to a steering command via a mapping function f(κ) with a lead filter for stable control. Figure 1 illustrates this process and the influence of R on clothoid generation. Lookahead distance R strongly affects tracking behavior: short R values can cause excessive steering and oscillations, while long R values may degrade tracking of tight curves at high speeds. As shown in Figure 1b, the controller's performance is sensitive to R, motivating the dynamic adaptation strategies introduced in this work to improve tracking accuracy and stability under varying conditions.

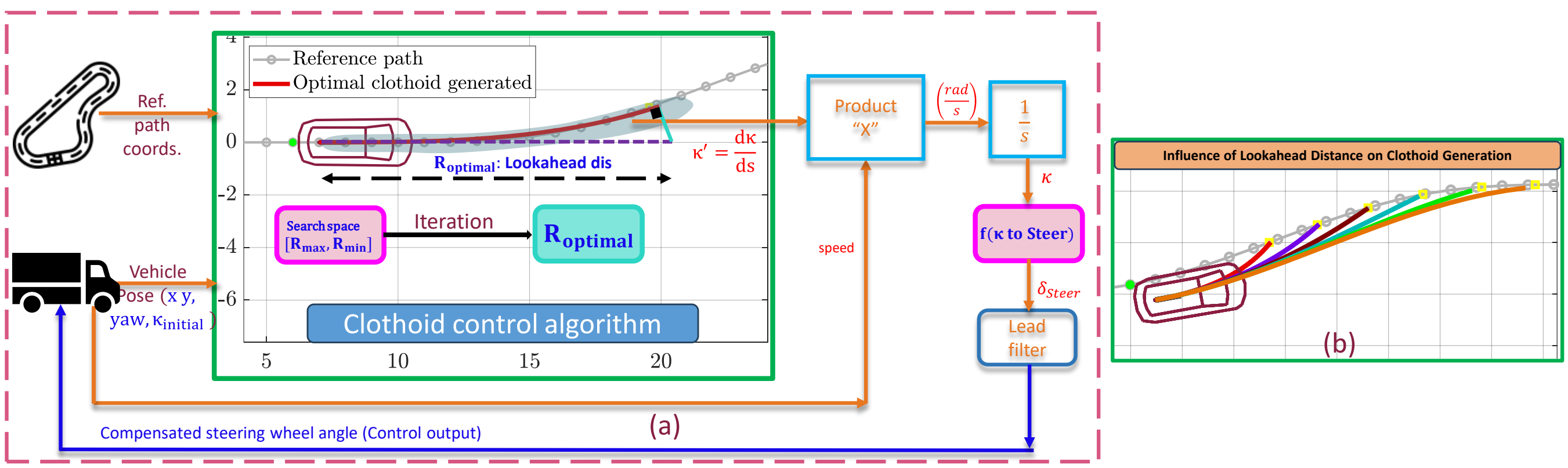

Figure 1 Overview of the clothoid controller framework: (a) Iterative clothoid generation process with optimal lookahead determination and control output, (b) influence of lookahead distance on clothoid path shaping.

## 2.2 Core Enhancements for High-Speed Operation

To improve performance at high speeds, particularly in managing oscillations and adapting to varying path geometries, several key enhancements to the lookahead determination process were developed. These improvements address the limitations of a fixed lookahead distance and help the controller respond more effectively to sharp curvature changes and transient disturbances. The methods are summarized below, with further implementation details provided in [11]. Figure 2 illustrates key aspects of these enhancements.

### a. Oscillation Damping via Integrated Tangential Check Criteria

A significant challenge at high speeds, especially during transitions from curved to straight path segments, is the tendency for oscillations even when the cross-track error (CTE) is relatively small. This occurs when the fitted clothoid trajectory does not align tangentially with the reference path at the lookahead point, leading to repeated overcorrections. To address this, a tangential check was introduced to the clothoid generation algorithm. This check evaluates whether the endpoint tangent angle of the candidate clothoid aligns smoothly with the reference path tangent at the lookahead point. By rejecting clothoids that introduce excessive heading mismatch, the method prevents abrupt curvature shifts and produces smoother steering commands during transitions, thereby improving overall path stability (Figure 2a).

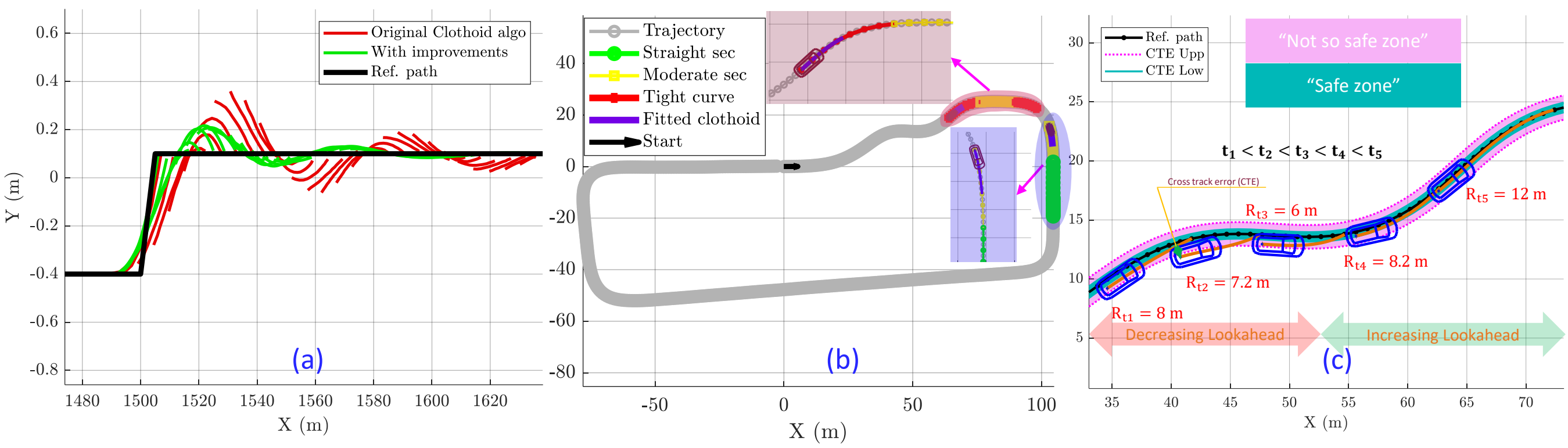

Figure 2 Illustrative examples of key high-speed enhancements: (a) Oscillation damping of generated clothoids at each time steps through tangential check, (b) dynamic lookahead range based on segment classification, and (c) dual-adaptive lookahead rate control responding to CTE and path changes.

### b. Dynamic Lookahead Range via Trajectory Segment Classification

A fixed lookahead range is often unsuitable for varying path geometries. To enable more context-aware lookahead selection, a real-time trajectory segment classification system was developed. Using polynomial-based curvature estimation, each segment of the reference path is classified as straight, moderate curve, or sharp curve. Based on this classification, the allowable range for the lookahead distance is adjusted dynamically: longer lookaheads are favored for straight segments to reduce unnecessary corrections, while shorter lookaheads are used in sharp curves to improve tracking accuracy. This adaptive range selection improves both steering smoothness and trajectory conformity across diverse driving conditions (Figure 2b).

Based on the classification, the lookahead distance range is adapted as follows:

- **Straight Segment**: $R_{max} = 13m$, $R_{min} = 11m$
- **Moderate Curve Segment**: $R_{max} = 10m$, $R_{min} = 8m$
- **Tight Curve Segment**: $R_{max} = 7m$, $R_{min} = 6m$

### c. Dual-Adaptive Lookahead Rate Control

In addition to adjusting the lookahead range, a dual-adaptive mechanism was implemented to control the rate of change of the lookahead distance. This mechanism combines two factors: (1) a cross-track error (CTE)-driven adjustment, where larger errors trigger faster reductions in lookahead to enable quicker path correction, and (2) a proactive segment-based adjustment, where the controller anticipates upcoming path changes (based on curvature classification) and gradually modifies the lookahead before encountering tight curves. This dual adaptation enables smoother transitions between different driving scenarios and prevents sudden steering inputs, further enhancing stability at high speeds (Figure 2c).

## 2.3 Extending the Framework for Articulated Vehicle Dynamics

The successful application of lateral control to articulated vehicles, such as semi-tractor trailers, requires addressing complexities not typically encountered with single-body vehicles. The core challenge stems from their multi-body structure, which leads to distinct dynamic behaviors that must be accounted for by the control system.

### a. The Challenge of Multiple Curvatures in Articulated Vehicles

Unlike single rigid-body vehicles where a single representative point (e.g., the center of the rear axle) can often adequately define the vehicle's path, articulated vehicles consist of two or more linked units (e.g., a tractor and a semitrailer). During maneuvers, particularly on curved sections of a path, these units do not follow a single, unified trajectory. Instead, the tractor and the trailer will each trace distinct paths with different radii of curvature, as illustrated in Figure 3. This phenomenon, known as off-tracking, becomes more pronounced in tighter curves. Consequently, a lateral controller designed with a single-body assumption will struggle to accurately align the entire articulated combination with a reference trajectory. Simply controlling one point on the tractor, for instance, may lead to significant deviations of the trailer from the intended path, or vice-versa, impacting safety, efficiency, and potentially causing encroachment into adjacent lanes or contact with obstacles. Therefore, an effective lateral control framework for articulated vehicles must acknowledge and manage these multiple, distinct curvatures.

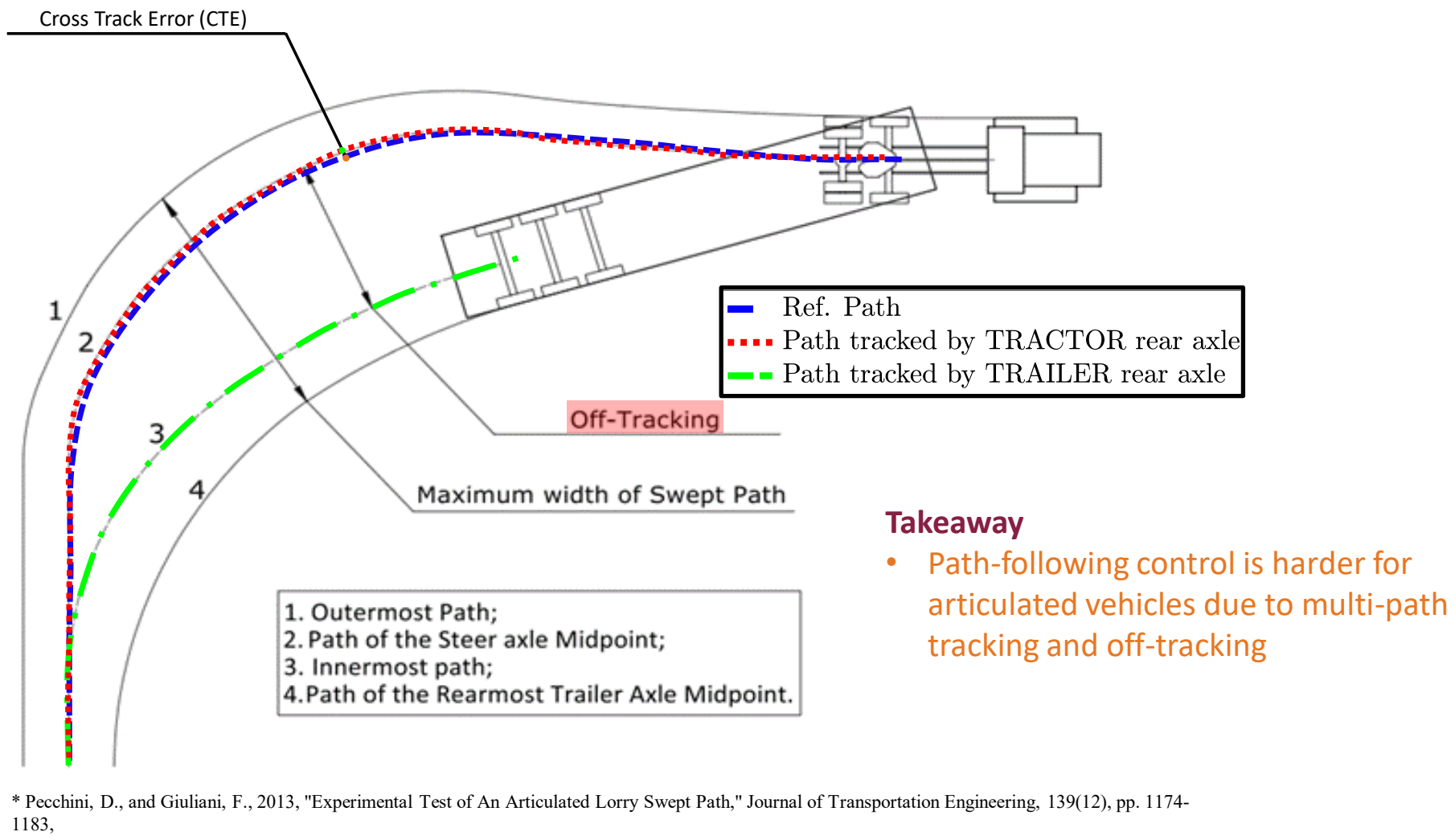

Figure 3 Off-tracking and distinct path curvatures of tractor and trailer units in an articulated vehicle during a turn.

### b. Implementing Curvature Control via Lookup Tables for Arbitrary Tracking Points

While a simple kinematic relationship, such as one derived from Ackermann steering geometry, can define the $f(\kappa \text{ to Steer})$ function to map a desired curvature ($\kappa$) at the tractor's rear axle to a steering angle ($\delta_{steer}$), this approach becomes inadequate for articulated vehicles when the control objective is to track other points (e.g., the steer axle, hitch point, or trailer center of gravity (CG)). Deriving analytical kinematic or dynamic relationships for the curvature at these arbitrary points as a direct function of steering input is significantly more complex, often impractical for real-time control, and highly sensitive to vehicle parameters and articulation angles.

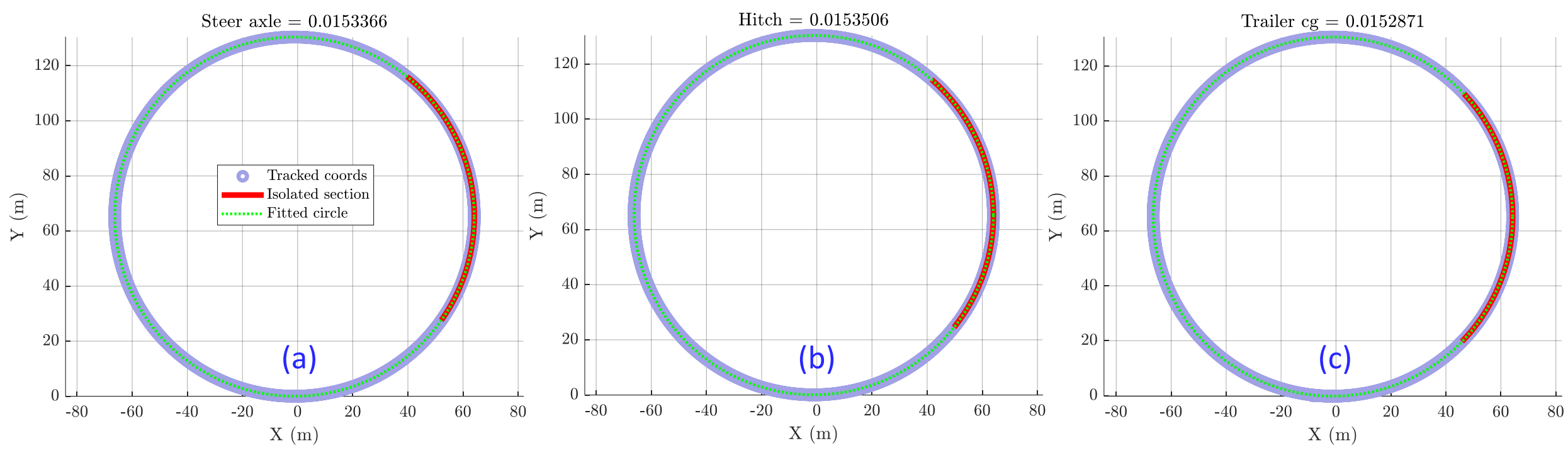

Figure 4 Curvature Estimation via Best Circle Fitting for Tracking Points. (a) Steering axle, (b) Hitch point, and (c) Trailer CG

To overcome this limitation and enable precise curvature control at various strategic points on an articulated vehicle, a lookup table (LUT) based approach was developed. This method empirically establishes the steady-state relationship between the steering input and the resulting path curvature at selected tracking points. The LUTs are generated offline through a series of high-fidelity simulations (e.g., using TruckSim®). In these simulations, the vehicle is driven at various constant speeds while a range of constant steering angles is applied. For each speed-steering combination, once the vehicle reaches a steady-state turning condition, the path traced by the desired tracking point (e.g., steer axle, hitch, trailer CG) is recorded and used to estimate the curvature of the path. As detailed in [11], methods such as fitting a circle to the steady-state trajectory data (as illustrated Figure 4) or applying smoothened curvature evaluation to recorded waypoints were used, yielding consistent curvature values. This process is repeated across the operational envelope of speeds and steering angles to populate the LUTs.

The resulting tracking points that map the vehicle speed and steering angle to curvature for each tracking point are shown in Figure 5. For control implementation, they are used as a look-up table of the inverse maps of the vehicle speed and desired curvature at tracking point, to determine the required steering angle. During online operation, when the clothoid controller determines an optimal curvature ($\kappa'$) for the chosen tracking point on the articulated vehicle, this desired curvature, along with the current vehicle speed, is used to interpolate the necessary steering angle from the corresponding LUT. The LUT-based f(κ to Steer) mechanism, specific to the chosen tracking point, allows the core clothoid controller (which already includes a lead filter to compensate for the lateral dynamics) to effectively manage the path of different parts of the articulated vehicle, a necessity for accurate overall path following.

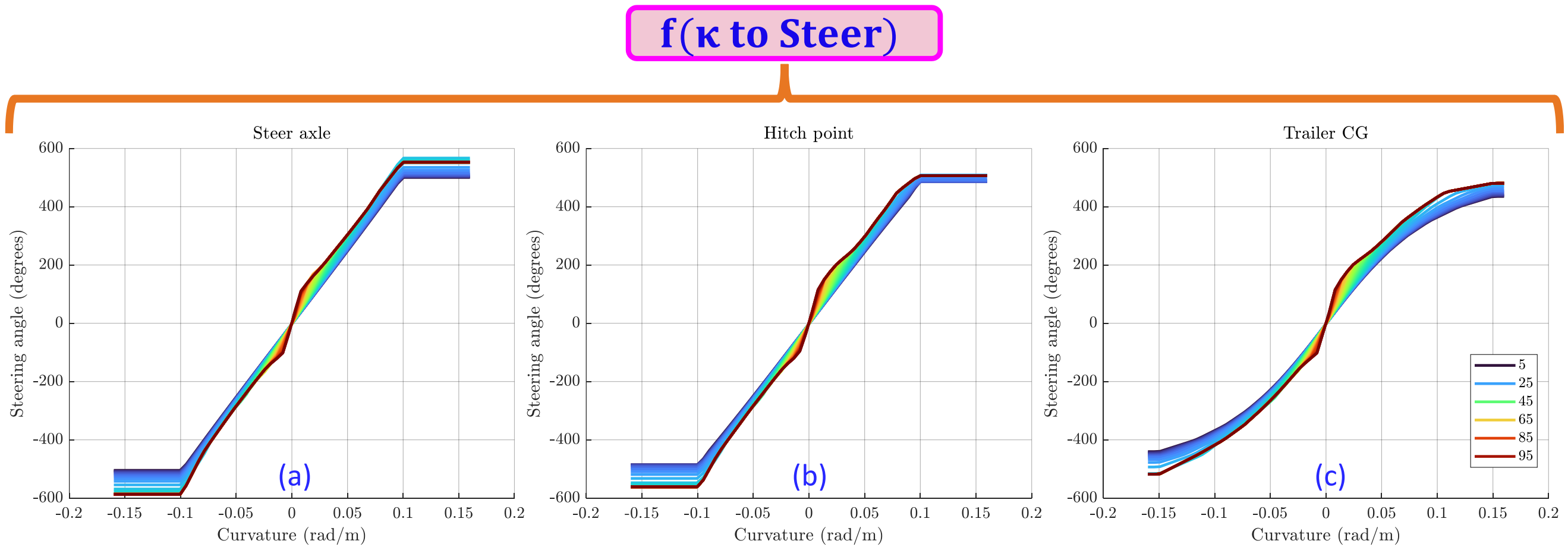

Figure 5 Steering Angle Vs Curvature for Different Tracking Points at Varying Speeds. (a) Steer Axle, (b) Hitch Point, (c) Trailer CG

### c. Enabling Scalability through Flexible Tracking Point Selection

The selection of the control point on an articulated vehicle significantly impacts lateral control performance due to the distinct paths traced by the tractor and trailer. To determine an optimal tracking strategy for the subsequent high-speed evaluations in this paper, an extensive preliminary study (detailed in [11], Section 5.6.1]) was conducted. This study analyzed Cross-Track Errors (CTEs) for various candidate tracking points positioned uniformly along the longitudinal centerline of the tractor, from the steer axle (defined as the origin in the tractor's local frame) to the rear axle (located 3.5 meters behind the steer axle). Performance was evaluated across challenging scenarios, including a high-speed Dual Lane Change (DLC) maneuver (refer Figure 6) and a figure-eight trajectory with sharp curvatures, considering individual CTEs of the tractor and trailer.

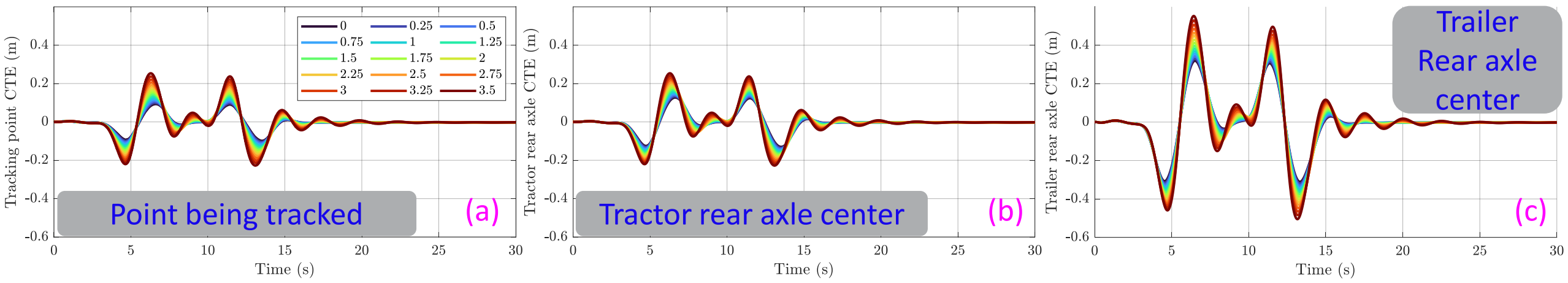

Figure 6 CTE analysis for different tracking points along the tractor centerline during a DLC maneuver, showing (a) tracking point CTE, (b) tractor rear axle CTE, (c) trailer rear axle CTE

The findings of this evaluation indicated that while different tracking points offered advantages for specific CTE components or maneuvers (e.g., the fifth wheel/hitch point sometimes yielded lower tractor CTE), tracking the steer axle center consistently provided the best overall balance, as shown in Figure 7. It yielded minimal combined cross-track error across the diverse scenarios and demonstrated superior performance in minimizing oscillations and maintaining stability, particularly during high-speed transitions. Based on this comprehensive analysis, the steer axle of the tractor was identified as the optimal tracking point for the controller in this study. Consequently, all subsequent simulations and results presented in this paper focus on tracking the steer axle to ensure robust and optimal path-following performance for the articulated vehicle configurations. This approach, combined with the lookup table-based curvature control, allows the framework to effectively scale from single-body vehicles to complex articulated systems.

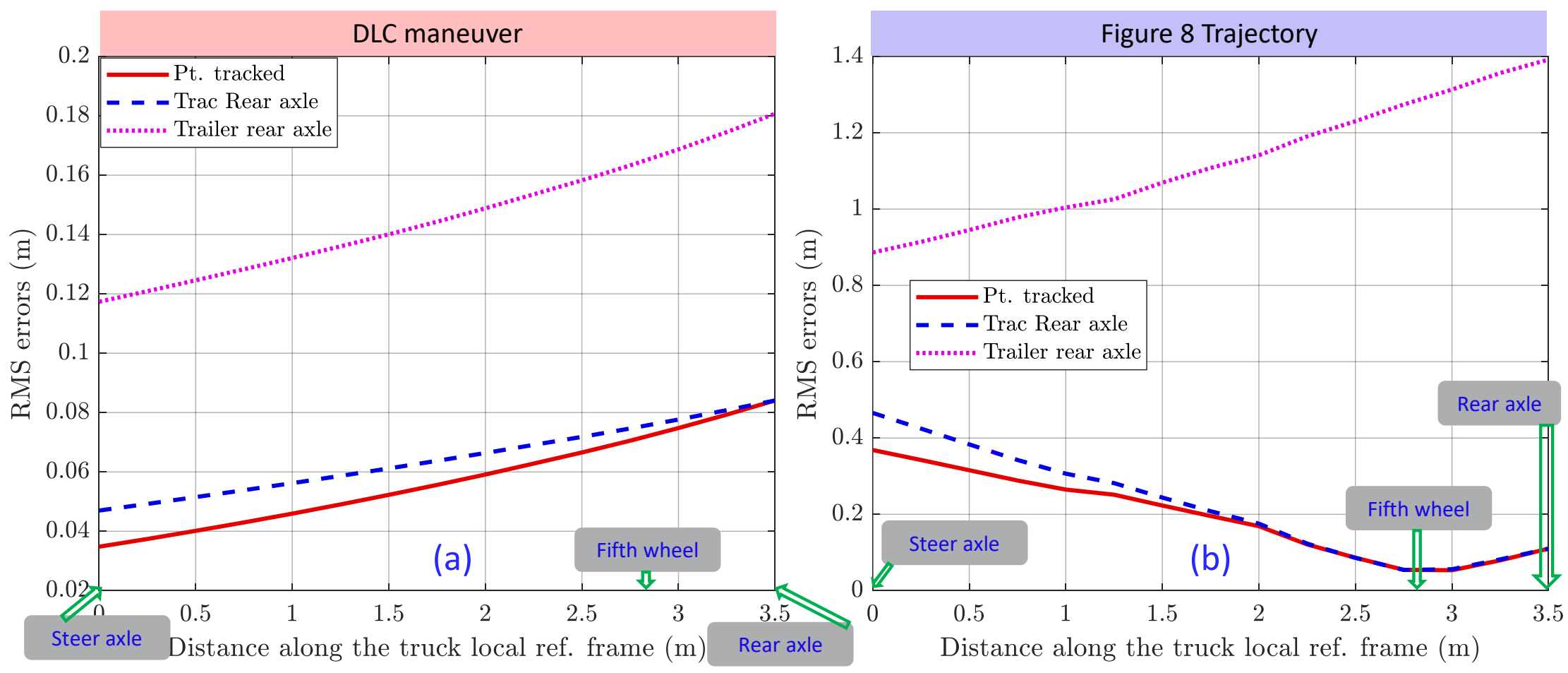

Figure 7 Summary of RMS CTE for various tracking points during (a) DLC maneuver and (b) figure-eight trajectory

# 3. Results and Discussions

The developed lateral controller was validated through co-simulation using MATLAB/Simulink® for the controller implementation and TruckSim® for high-fidelity vehicle dynamics modeling. Two primary vehicle configurations were employed to assess the controller's performance and scalability. The first configuration represented a single-body vehicle: a standalone 2-axle day cab tractor equipped with a 225 kW (approximately 300 HP) engine. The second configuration, designed to test scalability to more complex dynamics, consisted of the same 2-axle day cab tractor coupled with a 10-meter (approximately 33-foot) single-axle semitrailer carrying a payload of 4000 kg. All simulations were conducted on a surface with an asphalt friction coefficient of 0.85. Communication between TruckSim® and Simulink® occurred at a frequency of 1000 Hz to ensure numerical stability within the TruckSim® environment; however, the lateral controller itself operated at a more realistic sampling frequency of 100 Hz, reflecting common industry practice for such control systems.

## 3.1 Performance on Shasta Lake Section (I-5, California)

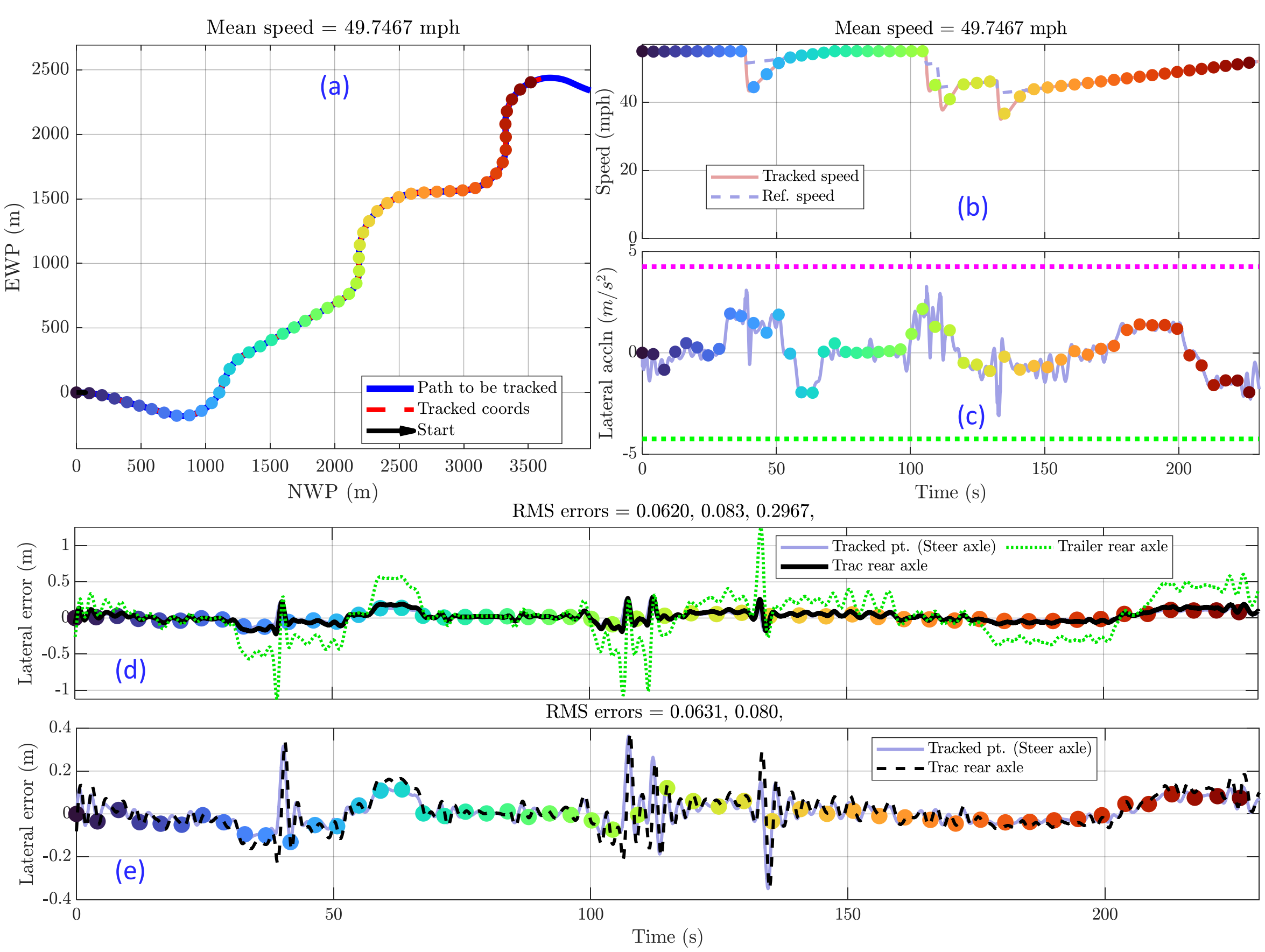

Figure 8 Path tracking results on the Shasta Lake I-5 section. (a) Bird's eye view of tracked path, (b) speed profile, (c) lateral acceleration, (d) lateral errors for tractor-trailer combination, (e) lateral errors for standalone tractor.

Figure 8 summarizes the controller's performance on the Shasta Lake section of I-5 in California, USA, which includes moderately curved segments with gradual curvature changes. The path geometry and color-coded progression are shown in Figure 8a; It is to be noted that the colored markers (●, ●,…), spaced at 200-meter intervals, serve as spatial references throughout each simulation. These markers are instrumental in correlating spatial data with time-domain results such as Speed (Figure 8b), cross-track errors (CTEs) (Figure 8d, e), and lateral acceleration (Figure 8c). For the articulated tractor-trailer combination, the controller maintained low cross-track errors (CTEs) for the tracked steer axle and tractor rear axle, with RMS values of 0.0620 m and 0.083 m, respectively. The trailer rear axle exhibited larger deviations, as expected from off-tracking effects in articulated vehicles, with an RMS error of 0.2967 m and peaks generally within ±1.2 m. Higher errors were observed around t = 50 s and 100–125 s, corresponding to sharper curves. The vehicle followed the target speed closely, while lateral acceleration remained comfortably within ±0.5 g, ensuring stable and smooth handling throughout the maneuver.

These results highlight the controller's ability to manage both path tracking and dynamic stability effectively at highway speeds.

When tested on the same route with a standalone tractor, the controller again achieved excellent tracking performance, with RMS CTEs of 0.0631 m and 0.080 m for the steer axle and tractor rear axle, respectively. The similarity of these results to those of the tractor-trailer combination illustrates the controller's scalability across different vehicle configurations.

## 3.2 Performance on Cabbage Hill Section (I-84, Oregon)

The controller's robustness was further tested on a notoriously challenging section of Interstate 84, Cabbage Hill in Oregon, USA, characterized by steep grades, multiple sharp turns, as depicted by the path in Figure 9. Simulations for both vehicle configurations maintained a mean speed of 75 km/hr. (46.6 mph). The maximum speed was limited to 89 km/hr. (55 mph), the posted speed limit for this section of the interstate.

For the articulated tractor-trailer, the steer axle and tractor rear axle maintained low CTEs with RMS values of 0.0508 m and 0.076 m, respectively, while the trailer rear axle exhibited higher deviations (RMS 0.3393 m), particularly in the sharper curved segments between t = 100–200 s and t = 300–400 s. These regions correspond to the steepest hairpin sections of the route. Despite this, trailer oscillations remained controlled, with lateral errors generally contained within ±1.0 m, demonstrating effective management of trailer dynamics under demanding conditions. The vehicle tracked speed accurately across the variable gradients, and lateral accelerations remained within safe margins (under ±0.5 g), confirming robust handling and ride stability.

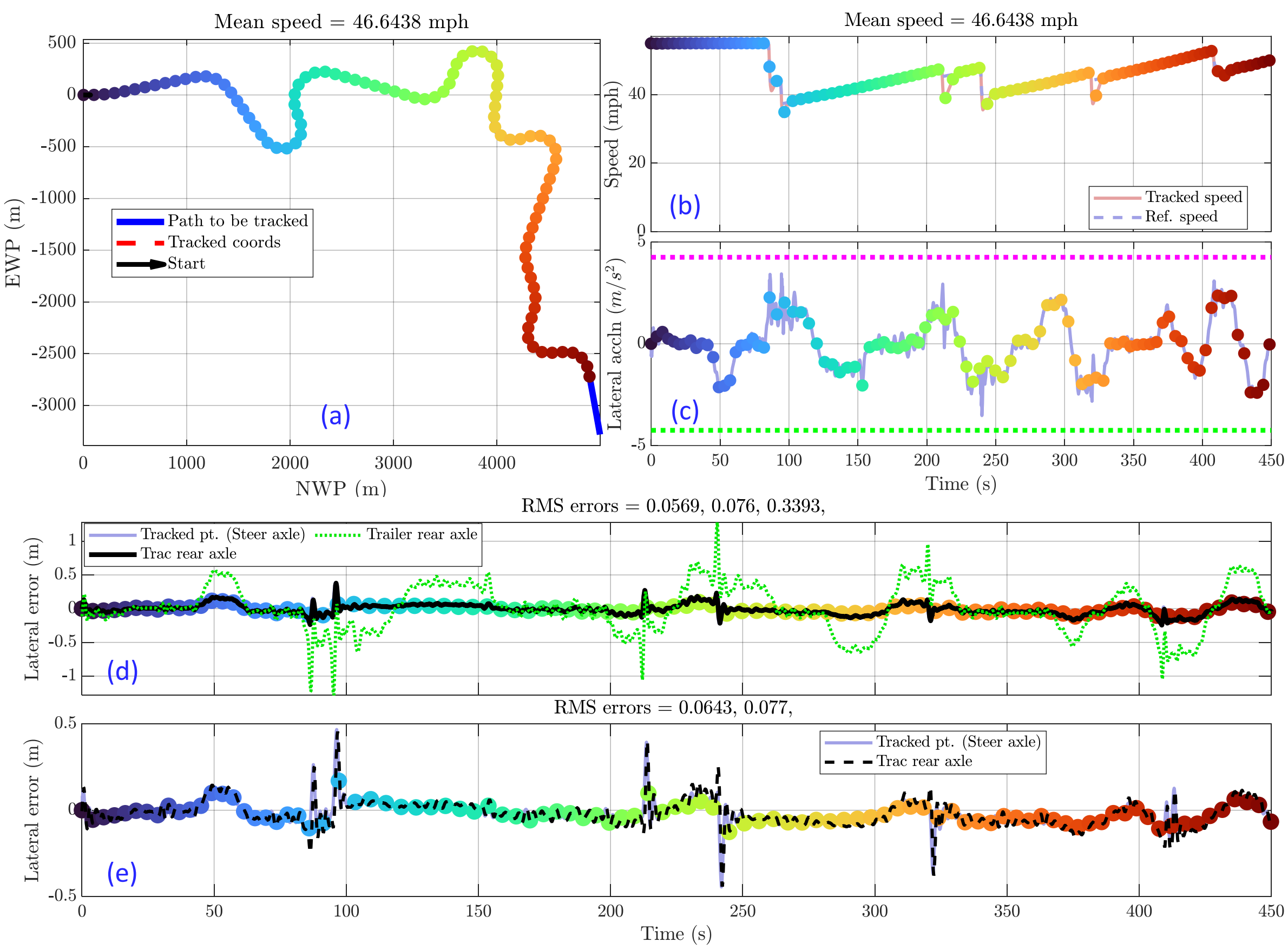

Figure 9 Lateral tracking performance on Cabbage Hill section of I-84. (a) Tracked path; (b) Speed profile; (c) Lateral acceleration; (d) Cross-track errors for articulated vehicle; (e) Cross-track errors for standalone tractor.

For the standalone tractor, which represents a straight vehicle, the controller again achieved strong performance, with steer axle and tractor rear axle RMS CTEs of 0.0643 m and 0.077 m. The sharp peaks in the standalone tractor's CTE plot (e.g., around t=90s, t=240s) correspond to the abrupt directional changes in the hairpin-like turns, yet the controller quickly recovers.

The Cabbage Hill results highlight the controller's ability to handle highly challenging real-world road geometries at considerable speeds for both single-body and articulated vehicles. The adaptive lookahead strategy and robust clothoid fitting allowed for effective negotiation of tight turns while maintaining overall stability and path adherence for the controlled point.

# 4. Conclusions and Future Works

A scalable high-speed lateral control method for autonomous vehicles was developed, using a clothoid-based architecture with key adaptations for enhanced performance across both single-body and articulated vehicle configurations. The proposed control strategy leverages adaptive lookahead mechanisms combining trajectory segment classification, tangential checks, and dual-adaptive rate control to mitigate oscillations, improve path tracking accuracy, and maintain stability under high-speed and high-curvature conditions. Additionally, the development of lookup table-based curvature control and flexible tracking point selection enables seamless extension of the method to articulated vehicles, accommodating their off tracking and cross track characteristics.

Extensive high-fidelity simulations were conducted on two actual highways in the USA, a moderately curved and steep section of Shasta Lake in California, USA and a winding section of Cabbage Hill. They were used to evaluate the controller's lateral path tracking effectiveness and robustness. The results demonstrate consistently low cross-track errors, smooth lateral acceleration profiles, and effective management of trailer dynamics at speeds up to 89 km/hr. (55 mph), the roadway's speed limit. The results further confirm the scalability and adaptability of the controller.

Future work will focus on further extending the approach to more complex articulated configurations, such as trucks with two or more trailers and validating the controller performance through simulation and possibly road testing.

# 5. References

[1] Shaju A, Southward S, Ahmadian M. Enhancing Autonomous Vehicle Navigation with a Clothoid-Based Lateral Controller. Applied Sciences (Switzerland) 2024;14. https://doi.org/10.3390/app14051817.

[2] Lal DS, Vivek A, Selvaraj G. Lateral control of an autonomous vehicle based on Pure Pursuit algorithm. Proceedings of 2017 IEEE International Conference on Technological Advancements in Power and Energy: Exploring Energy Solutions for an Intelligent Power Grid, TAP Energy 2017 2018:1–8. https://doi.org/10.1109/TAPENERGY.2017.8397361.

[3] Li S, Li Z, Zhang B, Zheng S, Lu X, Yu Z. Path tracking for autonomous vehicles based on nonlinear model: Predictive control method. SAE Technical Papers, vol. 2019- April, SAE International; 2019. https://doi.org/10.4271/2019-01-1017.

[4] Lukassek M, Dahlmann J, Völz A, Graichen K. Model predictive path-following control for truck–trailer systems with specific guidance points — design and experimental validation. Mechatronics 2024;100. https://doi.org/10.1016/j.mechatronics.2024.103190.

[5] Matute J, Diaz S, Karimoddini A. Sliding Mode Control for Robust Path Tracking of Automated Vehicles in Rural Environments. IEEE Open Journal of Vehicular Technology 2024:1–11. https://doi.org/10.1109/ojvt.2024.3456035.

[6] Garrow AL, Peters DL, Panchal TA. Curvature Sensitive Modification of Pure Pursuit Control. ASME Letters in Dynamic Systems and Control 2024;4. https://doi.org/10.1115/1.4064515.

[7] Ahn J, Shin S, Kim M, Park J. Accurate Path Tracking by Adjusting Look-Ahead Point in Pure Pursuit Method. International Journal of Automotive Technology 2021;22:119–29. https://doi.org/10.1007/s12239-021-0013-7.

[8] Chen C, Tomizuka M. Lateral Control of Tractor-Semitrailer Vehicles in Automated Highway Systems 1996.

[9] Jujnovich BA, Cebon D. Path-following steering control for articulated vehicles. Journal of Dynamic Systems, Measurement and Control, Transactions of the ASME 2013;135. https://doi.org/10.1115/1.4023396.

[10] Chen J, Jiang S, Zhou Z, Zhang M, Ming X, Guo N. Lateral semi-trailer truck control using a parameter self-learning MPC method in urban environment. Proceedings of the Institution of Mechanical Engineers, Part D: Journal of Automobile Engineering 2024;238:964–76. https://doi.org/10.1177/09544070221149068.

[11] Shaju A. Adaptive Longitudinal and Lateral Control for Autonomous Vehicles: High-Speed Platooning of Articulated Trucks 2024.